\documentclass{article}

\usepackage[final,nonatbib]{nips_2016}

\usepackage{graphicx}
\usepackage{amsmath}

\title{Learning binary or real-valued time-series via spike-timing dependent plasticity}

\author{
  Takayuki Osogami\\
  IBM Research - Tokyo\\
  \texttt{osogami@jp.ibm.com}
}

\begin{document}

\maketitle

\begin{abstract} A dynamic Boltzmann machine (DyBM) has been proposed as 
a model of a spiking neural network, and its learning rule of maximizing 
the log-likelihood of given time-series has been shown to exhibit key 
properties of spike-timing dependent plasticity (STDP), which had been 
postulated and experimentally confirmed in the field of neuroscience as 
a learning rule that refines the Hebbian rule.  Here, we relax some of 
the constraints in the DyBM in a way that it becomes more suitable for 
computation and learning. We show that learning the DyBM can be 
considered as logistic regression for binary-valued time-series.  We 
also show how the DyBM can learn real-valued data in the form of a 
Gaussian DyBM and discuss its relation to the vector autoregressive (VAR) 
model.  The Gaussian DyBM extends the VAR by using additional 
explanatory variables, which correspond to the eligibility traces of the 
DyBM and capture long term dependency of the time-series. Numerical 
experiments show that the Gaussian DyBM significantly improves the 
predictive accuracy over VAR.\end{abstract}

\section{Introduction}

The dynamic Boltzmann machine (DyBM) \cite{RT0967,DyBM} has recently 
been proposed as a model of a spiking neural network whose learning rule 
that maximizes the log likelihood of given {\em time-series} exhibits 
key properties of spike-timing dependent plasticity (STDP).  In STDP, 
the amount of the change in the synaptic strength between two neurons 
that fired together depends on the precise timings when the two neurons 
fired.  STDP supplements the Hebbian rule \cite{Hebb} and has been 
experimentally confirmed in biological neural networks \cite{BiPoo98}.  
Although the basic capability of the DyBM in learning time-series has 
been demonstrated in \cite{DyBM}, its application has been limited to 
relatively simple tasks with low dimensional and binary-valued 
time-series data.

Here, we relax some of the constraints that the DyBM has required in 
\cite{RT0967,DyBM} in a way that it becomes more suitable for 
computation and learning.  The primary purpose of these constraints in 
\cite{RT0967,DyBM} was to mimic a particular form of STDP.  Our 
relaxed DyBM generalizes the original DyBM and allows us to interpret 
it as a form of logistic regression for time-series data.  

We also discuss how the DyBM can deal with real-valued time-series in the form 
of a Gaussian DyBM, which is analogous to how Gaussian Boltzmann 
machines \cite{GaussianBM,WRH04,HinSal06} deal with real-valued patterns 
as opposed to Boltzmann machines \cite{AHS85,HinSej83} for binary values. 
Our Gaussian DyBM can be related to a vector autoregressive (VAR) model.
Specifically, we show that a special case of the Gaussian DyBM is a VAR
model having additional variables that capture long term dependency of
time-series.  These additional variables correspond to DyBM's
eligibility traces, which represent how recently and frequently spikes
arrived from a neuron to another.  

In addition, we  demonstrate the effectiveness of the Gaussian DyBM through
numerical experiments.  We train the Gaussian DyBM and let it predict the
future values of the time-series in a purely online manner with a
stochastic gradient method \cite{AdaGrad}.  Namely, at each moment, we
update the parameters and the variables of the Gaussian DyBM by using
only the latest values of the time-series, and let the DyBM predict the
next values of time-series.  The Gaussian DyBM can also be trained in a
distributed manner in that each parameter can be updated using only the
information that is locally available around the unit associated with
that parameter.  The experimental results show that the Gaussian DyBM
can reduce the predictive error by up to 20~\% against the
corresponding VAR without noticeably increasing computational cost.

The primary contribution of this paper is in the way that we relax the 
constraints in the original DyBM.  This relaxation allows us to 
represent the energy of the DyBM in a simple expression with matrices 
and vectors. Because the form of the energy completely determines the 
dynamics of the DyBM, our expression allows us to understand how the 
DyBM, a model of a spiking neural network, learns binary-valued 
time-series in a form of logistic regression.  The relaxation also 
allows us to relate the Gaussian DyBM to VAR.

\subsection{Related work}

There has been a significant amount of the prior work towards
understanding STDP from the perspectives of machine learning \cite{NPBM13,BMFZW16,SceBen16}.  For
example, Nessler et al.\ show that STDP can be understood as
approximating the expectation maximization (EM) algorithm
\cite{NPBM13}.  Nessler et al.\ study a particularly structured
(winner-take-all) network and its learning rule for maximizing
the log likelihood of given static patterns.  On the other hand, the
DyBM and the Gaussian DyBM do not assume particular structures in the
network, and the learning rule having the properties of STDP applies
for any synapse in the network.  Also, the learning rule of the DyBM
and the Gaussian DyBM maximizes the log likelihood of given
time-series, and its learning rule does not involve approximations
beyond what is assumed in stochastic gradient methods.

\section{Extending the dynamic Boltzmann machine}
\label{sec:DyBM}

We start by reviewing the DyBM as 
well as its learning rule that exhibits the key properties of STDP.  We 
then relax some of the constraints of the DyBM so that it has more 
flexibility in performing computation and learning time-series in a form 
of logistic regression.

\subsection{The dynamic Boltzmann machine}
\label{sec:DyBM:DyBM}

\begin{figure}
 \centering
 \includegraphics[width=\linewidth]{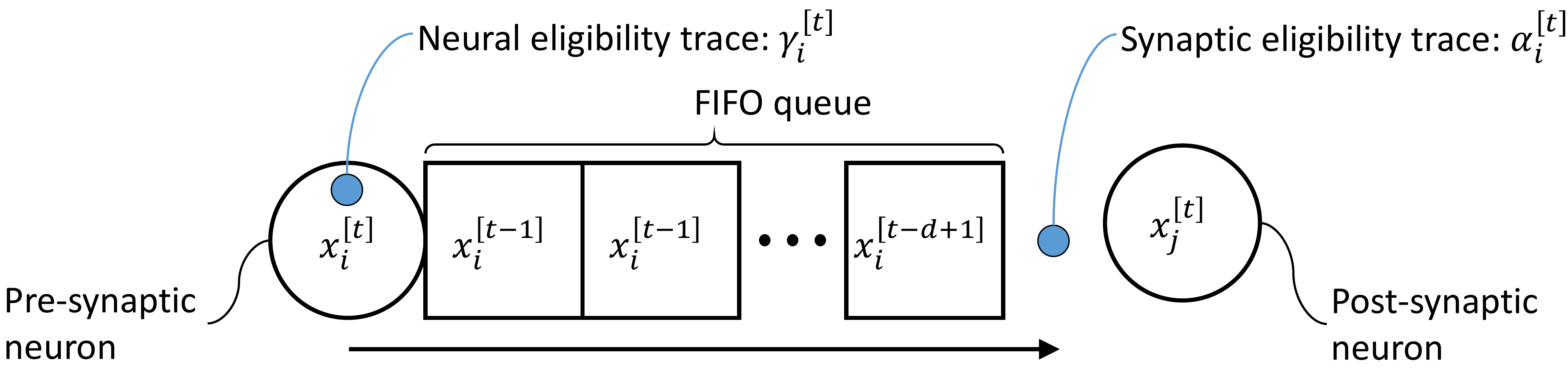}
 \label{fig:DyBM}
 \caption{A connection from a (pre-synaptic) neuron $i$ to a (post-synaptic) neuron $j$ in a DyBM.}
\end{figure}

A DyBM is an abstract model of a spiking neural network, where a 
(pre-synaptic) neuron is connected to a (post-synaptic) neuron via a 
first-in-first-out (FIFO) queue and a synapse (see Figure~\ref{fig:DyBM}). 
At each discrete time $t$, a neuron $i$ either fires ($x_i^{[t]}=1$) or 
not ($x_i^{[t]}=0$). The spike travels along the FIFO queue and reaches 
the synapse after conduction delay\footnote{For simplicity, we assume 
that the conduction delay is uniform for all connections, as opposed to 
variable conduction delay in \cite{DyBM}.  See also \cite{DelayPruning,HyperDyBM} for ways
to tune the values of the conduction delay.}, $d$.  In other words, the 
FIFO queue has the length of $d-1$ and stores, at time $t$, the spikes that have been 
generated by the pre-synaptic neuron from time $t-d+1$ to time $t-1$.

\begin{figure}
 \centering
 \includegraphics[width=0.5\linewidth]{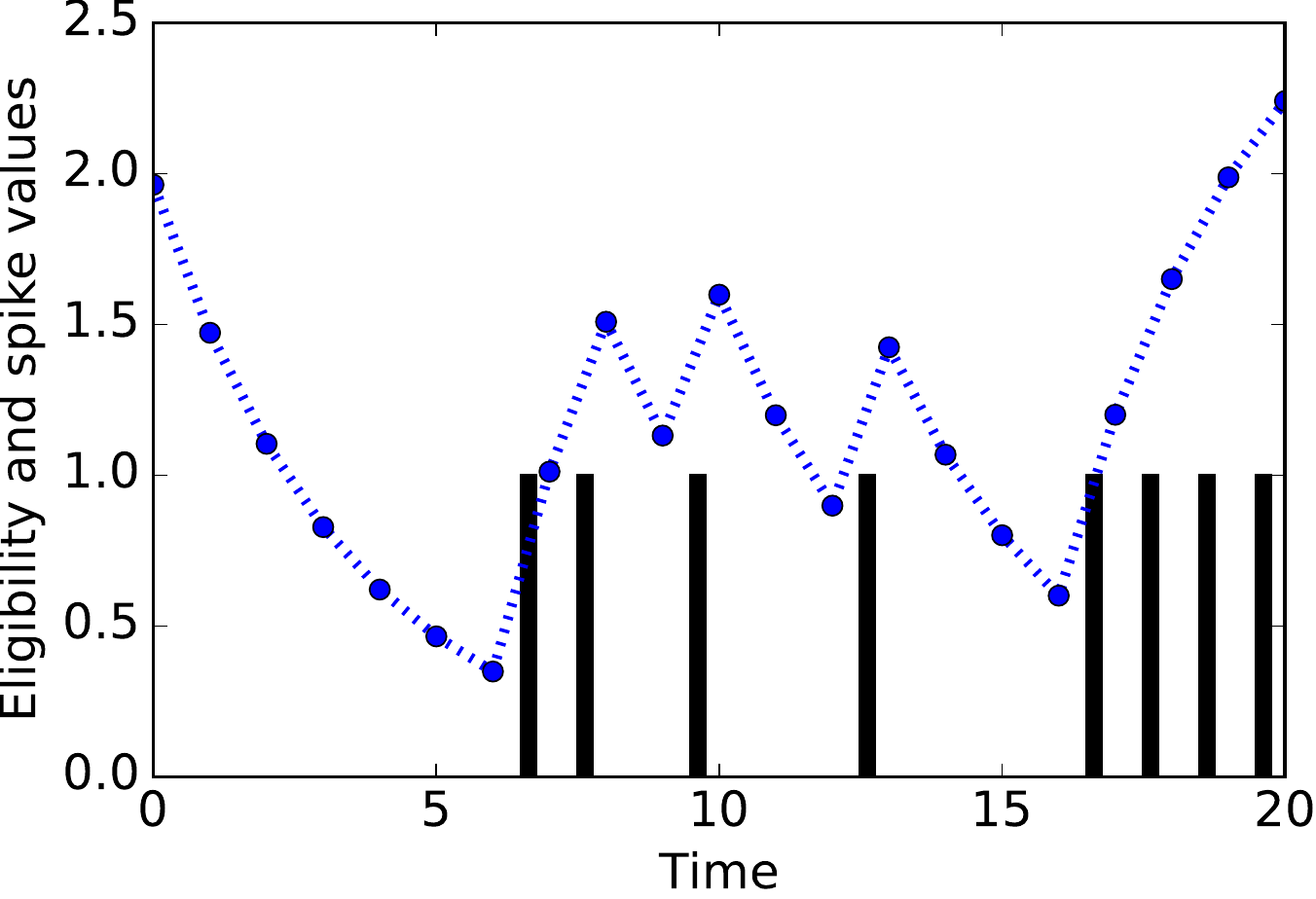}
 \label{fig:etrace}
 \caption{The value of a synaptic or neural eligibility trace as a function of time.
 For a synaptic eligibility trace at a synapse, the bars represent the spikes arrived from a FIFO queue at that synapse.
 For a neural eligibility trace at a neuron, the bars represent the 
 spikes generated by that neuron.}
\end{figure}

Each synapse in a DyBM stores a quantity called a synaptic 
eligibility trace\footnote{For simplicity, we assume a single synaptic eligibility trace, as opposed to 
multiple ones in \cite{DyBM}, at each synapse.}.  The value of the 
synaptic eligibility increases when a spike arrives at the synapse from the FIFO 
queue; otherwise, it is decreased by a constant factor.  
Specifically, at time $t$, the value of the synaptic eligibility trace,
$\alpha_{i}^{[t]}$, that is stored at the synapse from a pre-synaptic neuron $i$ is updated as 
follows:
\begin{align}
 \alpha_{i}^{[t]} = \lambda \, (\alpha_{i}^{[t-1]} + x_i^{[t-d+1]}),
 \label{eq:synaptic}
\end{align} 
where $\lambda$ is a decay rate and satisfies $0\le\lambda<1$.  
Figure~\ref{fig:etrace} shows an example of how the value of the synaptic 
eligibility trace changes depending on the spikes arrived at the synapse. 
 Observe that $\alpha_{i}^{[t]}$ represents how recently and 
frequently spikes arrived from a pre-synaptic neuron $i$ and can be represented non-recursively as follows:
\begin{align}
 \alpha_i^{[t-1]} = \sum_{s=-\infty}^{t-d} \lambda^{t-s-d} \, x_i^{[s]}.
\end{align}

Each neuron in a DyBM stores a quantity called a neural eligibility 
trace\footnote{We assume a single neural eligibility trace, as opposed 
to multiple ones in \cite{DyBM}, at each neuron.}. 
The value of the neural eligibility increases when the neuron fires; 
otherwise, it is decreased by a constant factor.  Specifically, at time 
$t$, the value of the neural eligibility trace, $\gamma_{i}^{[t]}$, at 
a neuron $i$ is updated as follows:
\begin{align}
 \gamma_{i}^{[t]} = \mu \, (\gamma_{i}^{[t-1]} + x_i^{[t]}),
 \label{eq:neural}
\end{align} 
where $\mu$ is a decay rate and satisfies $0\le\mu<1$.  
Observe that $\gamma_{i}^{[t]}$ represents how recently and 
frequently the neuron $i$ has fired and can be represented non-recursively as follows:
\begin{align}
 \gamma_{i}^{[t-1]} = \sum_{s=-\infty}^{t-1} \mu^{t-s} \, x_j^{[s]}
\end{align} 

A neuron in a DyBM fires according to the probability distribution that 
depends on the energy of the DyBM.  A neuron is more likely to fire when 
the energy becomes lower if it fires than otherwise.  
Let $E_j\big(x_j^{[t]}|\mathbf{x}^{[:t-1]}\big)$ be
the energy associated with a neuron $j$ at time $t$, which can depend on whether $j$ 
fires at time $t$ ({\it i.e.}, $x_j^{[t]}$) as well as the preceding spiking activities 
of the neurons in the DyBM ({\it i.e.}, $\mathbf{x}^{[:t-1]}$).  The firing probability of 
a neuron $j$ is then given by
\begin{align}
P_j( x_j^{[t]} | \mathbf{x}^{[:t-1]} )
& = 
\frac{\exp\big( -E_j(x_j^{[t]}|\mathbf{x}^{[:t-1]}) \big)}
{\displaystyle\sum_{\tilde x\in\{0,1\}} \exp\big(-E_j(\tilde x|\mathbf{x}^{[:t-1]})\big)}
\label{eq:probability}
\end{align}
for $x_j^{[t]}\in\{0,1\}$.
Specifically, $E_j\big(x_j^{[t]}|\mathbf{x}^{[:t-1]}\big)$ can be represented as follows:
\begin{align} 
E_j\big(x_j^{[t]}|\mathbf{x}^{[:t-1]}\big)
=
- b_j \, x_j^{[t]}
+ E_j^{\rm LTP} \big(x_j^{[t]}|\mathbf{x}^{[:t-1]}\big)
+ E_j^{\rm LTD} \big(x_j^{[t]}|\mathbf{x}^{[:t-1]}\big),
\label{eq:energy}
\end{align} 
where $b_j$ is the bias parameter of 
a neuron $j$ and represents how likely $j$ spikes ($j$ is more likely to 
fire if $b_j$ has a large positive value), and we define
\begin{align}
E_j^{\rm LTP} \big(x_j^{[t]}|\mathbf{x}^{[:t-1]}\big)
& \equiv
- \sum_{i} u_{i,j} \, \alpha_{i}^{[t-1]} \, x_j^{[t]} \label{eq:LTPenergy}\\
E_j^{\rm LTD} \big(x_j^{[t]}|\mathbf{x}^{[:t-1]}\big)
& \equiv 
\sum_{i} v_{i,j} \, \beta_{i}^{[t-1]} \, x_j^{[t]}
+ \sum_{k} v_{j,k} \, \gamma_{k}^{[t-1]} \, x_j^{[t]}, \label{eq:LTDenergy}
\end{align}
where $\beta_{i}^{[t-1]}$ represents how soon and frequently spikes will arrive
at the synapse from the FIFO queues from $i$ to $j$:
\begin{align}
\beta_{i}^{[t-1]} \equiv \sum_{s=t-d+1}^{t-1} \mu^{s-t} \, x_i^{[s]}.
\label{eq:beta}
\end{align}

In \eqref{eq:LTPenergy}, the summation with respect to $i$ is over all 
of the pre-synaptic neurons that are connected to $j$. Here, $u_{i,j}$ 
is the weight parameter from $i$ to $j$ and represents the strength of 
Long Term Potentiation (LTP).  This weight parameter is thus referred to 
as LTP weight.  A neuron $j$ is more likely to fire ($x_j^{[t]}=1$) when 
$\alpha_{i}^{[t-1]}$ is large for a pre-synaptic neuron $i$ connected to 
$j$ (spikes have recently arrived at $j$ from $i$) and the 
corresponding $u_{i,j}$ is positive and large (LTP from $i$ to $j$ is 
strong).

In \eqref{eq:LTDenergy}, the summation with respect to $i$ is over all 
of the pre-synaptic neurons that are connected to $j$, and the summation 
with respect to $k$ is over all of the post-synaptic neurons which $j$ 
is connected to. Here, $v_{i,j}$ represents the strength of Long Term 
Depression from $i$ to $j$ and referred to as LTD weight.  The neuron 
$j$ is less likely to fire when $\beta_{i}$ is large for a pre-synaptic 
neuron $i$ connected to $j$ (spikes will soon and frequently reach $j$ 
from $i$) and the corresponding $v_{i,j}$ is positive and large (LTD 
from $i$ to $j$ is strong).  The second term in \eqref{eq:LTDenergy} 
represents that a pre-synaptic neuron $j$ is less likely to fire if a 
post-synaptic neuron has recently and frequently fired ($\gamma_k$ is 
large), and the strength of this LTD is given by $v_{j,k}$.  Notice that 
the timing of a spike is measured with respect to when the spike reaches 
synapse, where the spike from a pre-synaptic neuron has the delay $d$, 
and the spike from a post-synaptic neuron reaches immediately.

The learning rule of the DyBM has been derived in a way that it maximizes the log likelihood of
given time-series with respect to the probability distribution given by \eqref{eq:probability} \cite{DyBM}.
Specifically, at time $t$, the DyBM updates its (plastic) parameters according to 
\begin{align}
b_j & \leftarrow b_j + \eta \, \big( x_j^{[t]} - \langle X_j^{[t]} \rangle \big) \label{eq:b} \\
u_{i,j} & \leftarrow u_{i,j} + \eta \, \alpha_{i}^{[t-1]} \, \big( x_j^{[t]} - \langle X_j^{[t]} \rangle \big) \label{eq:u} \\
v_{i,j} & \leftarrow v_{i,j} + \eta \, \beta_{i}^{[t-1]} \, \big( \langle X_j^{[t]} \rangle - x_j^{[t]} \big)
+ \eta \, \gamma_{j}^{[t-1]} \, \big( \langle X_i^{[t]} \rangle - x_i^{[t]} \big) \label{eq:v}
\end{align}
for each of neurons $i$ and $j$, where $\eta$ is a learning rate, 
$x_j^{[t]}$ is the training data given to $j$ at time $t$, and $\langle 
X_j^{[t]} \rangle$ denotes the expected value of $x_j^{[t]}$ ({\it i.e.}, 
firing probability of a neuron $j$ at time $t$) according to the 
probability distribution given by \eqref{eq:probability}.  

In \eqref{eq:b}, $b_j$ is increased when $x_j^{[t]}=1$ is given to $j$, 
so that $j$ becomes more likely to fire (in accordance with the training 
data), but the amount of the change in $b_j$ is small if $j$ is already 
likely to fire ($\langle X_j^{[t]}\rangle\approx 1$). This dependency on 
$\langle X_j^{[t]}\rangle$ can be considered as a form of homeostatic 
plasticity. 

In \eqref{eq:u}, $u_{i,j}$ is increased (LTP gets stronger) when 
$x_j^{[t]}=1$ is given to $j$.  Then $j$ becomes more likely to fire 
when spikes from $i$ have recently and frequently arrived at $j$ ({\it 
i.e.}, $\alpha_{i}^{[\cdot]}$ is large).  This amount of the change in 
$u_{i,j}$ depends on $\alpha_{i}^{[t-1]}$, exhibiting a key property of 
STDP.  In particular, $u_{i,j}$ is increased by a large amount if spikes 
from $i$ have recently and frequently arrived at $j$.  

According to the second term on the right-hand side of \eqref{eq:v}, 
$v_{i,j}$ is increased (LTD gets stronger) when $x_j^{[t]}=0$ is given 
to a post-synaptic neuron $j$.  Then $j$ becomes less likely to fire 
when spikes from $i$ are expected to reach $j$ soon ({\it i.e.}, 
$\beta_{i}^{[\cdot]}$ is large).  This amount of the change in $v_{i,j}$ 
is large if there are spikes in the FIFO queue from $i$ to $j$ and they 
are close to $j$. According to the last term of \eqref{eq:v}, $v_{i,j}$ 
is increased when $x_i^{[t]}=0$ is given to the pre-synaptic $i$, and 
this amount of the change in $v_{i,j}$ is proportional to $\gamma_j$ 
({\it i.e.}, how frequently and recently the post-synaptic $j$ has fired). 
This learning rule of \eqref{eq:v} thus exhibits some of the key properties of LTD 
with STDP.

\subsection{Giving flexibility to the DyBM}
\label{sec:DyBM:extend}

It has been shown in \cite{DyBM} that the DyBM in 
Section~\ref{sec:DyBM:DyBM} has the capability of associative memory and 
anomaly detection for sequential patterns, but the applications of the 
DyBM has been limited to simple tasks with relatively low dimensional 
time-series.  Here, we relax some of the constraints of this DyBM in a 
way that it gives more flexibility that is useful for learning and inference.

Specifically, observe that the first term on the right-hand side of \eqref{eq:LTDenergy} can be
rewritten with the definition of $\beta_{i}^{[t-1]}$ in \eqref{eq:beta} as follows:
\begin{align}
\sum_{i} v_{i,j} \, \beta_{i}^{[t-1]} \, x_j^{[t]}
& = \sum_{i} \sum_{s=t-d+1}^{t-1} v_{i,j} \, \mu^{s-t} \, x_i^{[s]} \, x_j^{[t]} \\
& = \sum_{i} \sum_{\delta=1}^{d-1} v_{i,j}^{[\delta]} \, x_i^{[t-\delta]} \, x_j^{[t]},
\end{align}
where we let $v_{i,j}^{[\delta]} \equiv v_{i,j} \, \mu^{-\delta}$.  
Here, $v_{i,j}^{[\delta]}$ represents how unlikely $j$ fires at time $t$ 
if $i$ fired at time $t-\delta$.  The parametric form of 
$v_{i,j}^{[\delta]} \equiv v_{i,j} \, \mu^{-\delta}$ assumes that
this LTD weight decays geometrically as the interval, $\delta$, between the two spikes increases.

In the following, we relax this constraint on $v_{i,j}^{[\delta]}$ for 
$\delta=1,\ldots,d-1$ and assumes that these LTD weights can take 
independent values.  Then the energy of the DyBM with $N$ neurons
can be represented conveniently with matrix and vector operations:
\begin{align} 
E(\mathbf{x}^{[t]}|\mathbf{x}^{[:t-1]})
& \equiv \sum_{j=1}^N E_j(x_j^{[t]}|\mathbf{x}^{[:t-1]}) \\
& =
- \mathbf{b}^\top \mathbf{x}^{[t]}
- (\boldsymbol{\alpha}_\lambda^{[t-1]})^\top \mathbf{U} \, \mathbf{x}^{[t]}
+ \sum_{\delta=1}^{d-1} (\mathbf{x}^{[t-\delta]})^\top \mathbf{V}^{[\delta]} \, \mathbf{x}^{[t]}
+ (\mathbf{x}^{[t]})^\top \mathbf{V} \, \boldsymbol{\gamma}_\mu^{[t-1]},
\label{eq:matrix}
\end{align}
where $\mathbf{b}\equiv(b_j)_{j=1,\ldots,N}$ is a 
vector, $\mathbf{U}\equiv(u_{i,j})_{(i,j)\in\{1,\ldots,N\}^2}$ is a 
matrix, and other boldface letters are defined analogously (a vector is 
lowercase and a matrix is uppercase).  
For eligibility traces ($\boldsymbol{\alpha}_\lambda^{[t-1]}$ and
$\boldsymbol{\gamma}_\mu^{[t-1]}$), we append the subscript to explicitly
represent the dependency on the decay rate ($\lambda$ and $\mu$). 
The functional form of the energy 
completely determines the dynamics of a DyBM, and relaxing its constraints 
allows the DyBM to represent a wider class of dynamical 
systems.

Notice that the last term of \eqref{eq:matrix} can be divided into two terms:
\begin{align}
(\mathbf{x}^{[t]})^\top \mathbf{V} \, \boldsymbol{\gamma}_\mu^{[t-1]}
& = (\boldsymbol{\gamma}_\mu^{[t-1]})^\top \mathbf{V} \, \mathbf{x}^{[t]} \\
& = (\boldsymbol{\alpha}_\mu^{[t-1]})^\top \mathbf{V} \, \mathbf{x}^{[t]}
+ \sum_{\delta=1}^{d-1} (\mathbf{x}^{[t-\delta]})^\top \mathbf{\hat V}^{[\delta]} \, \mathbf{x}^{[t]},
\label{eq:divide}
\end{align}
where $\boldsymbol{\alpha}_\mu^{[t-1]}$ is the same as the vector of 
synaptic eligibility traces but with the decay rate $\mu$, and 
$\mathbf{\hat V}^{[\delta]} \equiv \mu^{-\delta} \, \mathbf{V}$.
Comparing \eqref{eq:divide} and \eqref{eq:matrix}, we find that, without loss of generality, 
the energy of the DyBM
can be represented with the following form:
\begin{align} 
E(\mathbf{x}^{[t]}|\mathbf{x}^{[:t-1]})
& =
- \bigg(
	\mathbf{b}^\top
	+ \sum_{\delta=1}^{d-1} (\mathbf{x}^{[t-\delta]})^\top \mathbf{W}^{[\delta]}
	+ \sum_{\ell=1}^L (\boldsymbol{\alpha}_{\lambda_\ell}^{[t-1]})^\top \mathbf{U}_\ell
  \bigg) \, \mathbf{x}^{[t]},
\label{eq:general}
\end{align} 
where we define $\mathbf{W}^{[\delta]} = -\mathbf{V}^{[\delta]} - \mathbf{\hat V}^{[\delta]}$.
The energy in \eqref{eq:general} reduces to the original energy in \eqref{eq:energy} when
$\mathbf{W}^{[\delta]} = - \mu^{-\delta} \, \mathbf{V} - \mu^{\delta} \, \mathbf{V}^\top$,
$\mathbf{U}_1 = \mathbf{U}$,
$\mathbf{U}_2 = - \mu^d \, \mathbf{V}^\top$,
$\lambda_1    = \lambda$, 
$\lambda_2    = \mu$,
and $L=2$.  With $L>2$, one can also incorporate multiple synaptic or neural eligibility
traces with varying decay rates in \cite{DyBM}. 
Equivalently, we can represent the energy using neural eligibility traces, 
$\boldsymbol{\gamma}_{\mu_\ell}$, instead of synaptic eligibility traces,
$\boldsymbol{\alpha}_{\lambda_\ell}$, as follows:
\begin{align} 
E(\mathbf{x}^{[t]}|\mathbf{x}^{[:t-1]})
& =
- \bigg(
	\mathbf{b}^\top
	+ \sum_{\delta=1}^{d-1} (\mathbf{x}^{[t-\delta]})^\top \mathbf{W}^{[\delta]}
	+ \sum_{\ell=1}^L (\boldsymbol{\gamma}_{\mu_\ell}^{[t-1]})^\top \mathbf{V}_\ell
  \bigg) \, \mathbf{x}^{[t]}.
\label{eq:general2}
\end{align} 

\subsection{Logistic regression for time-series with the DyBM}

We now show that we are actually learning a kind of a logit model for time-series
by learning a DyBM.  Let
\begin{align}
\boldsymbol{m}^{[t]}
& \equiv \mathbf{b}^\top
	+ \sum_{\delta=1}^{d-1} (\mathbf{x}^{[t-\delta]})^\top \mathbf{W}^{[\delta]}
	+ \sum_{\ell=1}^L (\boldsymbol{\alpha}_{\lambda_\ell}^{[t-1]})^\top \mathbf{U}_\ell.
\label{eq:mean}
\end{align}
Then we can write \eqref{eq:energy} as $E_j(x_j^{[t]}|\mathbf{x}^{[:t-1]}) = - m_j^{[t]} \, x_j^{[t]}$.

The firing probability in \eqref{eq:energy} can now be expressed as
\begin{align}
P_j(x_j^{[t]}|\mathbf{x}^{[:t-1]})
& = \frac{\exp(m_j^{[t]} \, x_j^{[t]})}{1 + \exp(m_j^{[t]})}
\label{eq:logit}
\end{align}
for $x_j^{[t]}\in\{0,1\}$. Namely, $m_j^{[t]}$ represents the negative 
energy associated with a neuron $j$ on the condition that $j$ fires at 
time $t$, and $j$ is likely to fire at $t$ if $m_j^{[t]}$ is positive 
and large. Recall that $m_j^{[t]}$ depends on $\mathbf{x}^{[:t-1]}$. 

The form of \eqref{eq:logit} implies that the DyBM is a kind of a logit 
model, where the feature vector, $(\mathbf{x}^{[t-d+1]}, \ldots, 
\mathbf{x}^{[t-1]}, \boldsymbol{\alpha}_\lambda^{[t-1]}, 
\boldsymbol{\alpha}_\mu^{[t-1]})$, depends on the prior values, 
$\mathbf{x}^{[:t-1]}$, of the time-series. By applying the learning 
rules given in \eqref{eq:b}-\eqref{eq:v} to given time-series, we can 
learn the parameters of the DyBM or equivalently the parameters of the 
logit model ({\it i.e.}, $\mathbf{b}$, $\mathbf{W}^{[\delta]}$ for 
$\delta=1, \ldots, d-1$, and $\mathbf{U}_\ell$ for $\ell=1, \ldots, L$) in \eqref{eq:logit}.

\section{Gaussian dynamic Boltzmann machines}
\label{sec:GaussianDyBM}

In this section, we show how a DyBM can deal with real-valued 
time-series in the form of a Gaussian DyBM.  A Gaussian DyBM assumes 
that $x_j^{[t]}$ follows a Gaussian distribution for each $j$:
\begin{align}
 p_j(x_j^{[t]} | \mathbf{x}^{[t-T,t-1]})
 = \frac{1}{\sqrt{2\,\pi\,\sigma_j^2}}
 \exp\Big(-\frac{\big(x_j^{[t]}-m_j^{[t]}\big)^2}{2\,\sigma_j^2}\Big),
 \label{eq:spike_prob}
\end{align}
where $m_j^{[t]}$ is given by \eqref{eq:mean}, and $\sigma_j^2$ is a 
variance parameter.  This Gaussian distribution 
is in contrast to the Bernoulli distribution of the DyBM 
given by \eqref{eq:probability}.

We now derive a learning rule for the Gaussian DyBM in a way that it 
maximizes the log-likelihood of given time-series $\mathbf{x}^{[\cdot]}$:
\begin{align}
 \sum_t \log p(\mathbf{x}^{[t]}|\mathbf{x}^{[:t-1]})
& = \sum_t \sum_{i=1}^N \log p_i(x_i^{[t]}|\mathbf{x}^{[-\infty,t-1]}),
\label{eq:independence}
\end{align}
where the summation over $t$ is over all of the time steps of 
$\mathbf{x}^{[\cdot]}$, and the conditional independence between 
$x_i^{[t]}$ and $x_j^{[t]}$ for $i\neq j$ given $\mathbf{x}^{[:t-1]}$ is 
the fundamental property of the DyBM shown in \cite{DyBM}.

The approach of stochastic gradient is to update the parameters of the 
Gaussian DyBM at each step, $t$, according to the gradient of the 
conditional probability density of $\mathbf{x}^{[t]}$:
\begin{align}
 \nabla \log p(\mathbf{x}^{[t]}|\mathbf{x}^{[:t-1]})
& = - \sum_{i=1}^N
 \Big(\frac{1}{2} \nabla \log \sigma_i^2 + \nabla\frac{\big(x_i^{[t]}-m_i^{[t]})^2}{2\,\sigma_i^2}\Big),
\label{eq:gradient}
\end{align}
where the equality follow from
\eqref{eq:spike_prob}.
From \eqref{eq:gradient} and \eqref{eq:mean},
we can derive the derivative with respect to
each parameter.

These parameters are thus updated as follows\footnote{In Appendix~\ref{sec:natural},
we derive learning rules based on natural gradients \cite{NaturalGradient}.}:
\begin{align}
 b_j \leftarrow b_j + \eta \, \frac{x_j^{[t]}-m_j^{[t]}}{\sigma_j^2},
  & \hspace{10mm}
\sigma_j \leftarrow \sigma_j + \eta \, \Bigg( \frac{\big(x_j^{[t]}-m_j^{[t]}\big)^2}{\sigma_j^2} - 1 \Bigg) \, \frac{1}{\sigma_j},
\label{eq:learning_rule1}
 \\
 w_{i,j}^{[\delta]} \leftarrow w_{i,j}^{[\delta]} + \eta \, \frac{x_j^{[t]}-m_j^{[t]}}{\sigma_j^2} \, x_i^{[t-\delta]},
 & \hspace{10mm}
 u_{i,j,\ell} \leftarrow u_{i,j,\ell} + \eta \, \frac{x_j^{[t]}-m_j^{[t]}}{\sigma_j^2} \, \alpha_{i,\lambda_{\ell}}^{[t-1]}
\label{eq:learning_rule2}
\end{align}
for $\ell=1,\ldots,L$, $\delta=1,\ldots,d-1$, and 
$(i,j)\in\{1,\ldots,N\}^2$, where $\eta$ is the learning rate.  In 
\eqref{eq:learning_rule1}-\eqref{eq:learning_rule2}, $m_j^{[t]}$ is 
given by \eqref{eq:mean}, $u_{i,j,\ell}$ is the $(i,j)$ element of 
$\mathbf{U}_\ell$, and $\alpha_{i,\lambda_{\ell}}$ is the $i$-th element of 
$\boldsymbol{\alpha}_{\lambda_\ell}$.

The maximum likelihood estimator of $\mathbf{x}^{[t]}$ by the Gaussian DyBM
is given by $\boldsymbol{m}^{[t]}$ in $\eqref{eq:mean}$.  
The Gaussian DyBM can thus be understood as a modification 
to the standard VAR. Specifically, the last term in the right-hand side 
of \eqref{eq:mean} involves eligibility traces, which can be understood 
as features of historical values, $\mathbf{x}^{[:t-d]}$, and are added 
as new variables of the VAR model.  Because the value of the eligibility 
traces can depend on the infinite past, the Gaussian DyBM can take into 
account the history beyond the lag $d$.

\section{Numerical experiments}
\label{sec:exp}

We now demonstrate the advantages of the Gaussian DyBM through numerical 
experiments.  The purpose of our experiment is to demonstrate the 
effectiveness of the eligibility traces of the Gaussian DyBM.  
Specifically, we train the Gaussian DyBM with a one dimensional sequence, 
which is generated according to the following noisy sine 
wave:
\begin{align} 
 x^{[t]} = \sin(2\,\pi\,t/100) + \varepsilon^{[t]}
 \label{eq:sin}
\end{align}
for each $t$, where $\varepsilon^{[t]}$ is independent and identically
distributed with the standard Gaussian distribution.  All of the experiments are carried out with a Python 2.7 
implementation on a Linux machine having 32 cores of POWER8 and 64~GB 
memory.

We consider a Gaussian DyBM, with the representation \eqref{eq:general2}, 
having a single unit ($N=1$), which is connected to itself with a FIFO
queue of length $d$ and has a neural eligibility trace of decay rate $\mu$. 
We vary $d$ and $\mu$ in the experiment.  This Gaussian
DyBM makes a prediction, $m^{[t]}$, according to
\begin{align}
  m^{[t]} = b + \sum_{\delta=1}^{d-1} w^{[\delta]}\,x^{[t-\delta]} + v\,\gamma^{[t-1]},
  \label{eq:1Dmodel}
\end{align}
where $\gamma^{[t-1]}\equiv \sum_{s=1}^{\infty} \mu^{s-d} \,
x^{[t-s]}$, and ($b$, $w$, $v$) is the set of parameters of the Gaussian DyBM.
For $\mu=0$, we define
$\gamma^{[t-1]}=x^{[t-d]}$, and this Gaussian DyBM reduces to
a VAR model with $d$ lags.

We train the Gaussian DyBM in an online manner.  Namely, for each step
$t$, we give a pattern, $x^{[t]}$, to the Gaussian DyBM to
update its eligibility trace, FIFO queue, and
parameters, and then let the Gaussian DyBM predict the next pattern,
$x^{[t+1]}$.  This process is repeated sequentially for
$t=1,2,\ldots$.  Here, the parameters are updated according to natural
gradients \eqref{eq:natural_rule1}-\eqref{eq:natural_rule2}.  The
learning rate, $\eta$, in
\eqref{eq:natural_rule1}-\eqref{eq:natural_rule2} is adjusted for each
parameter according to AdaGrad \cite{AdaGrad}, where the initial
learning rate is set $\eta=0.001$.  Throughout, the initial values of
the parameters and variables, including eligibility traces and the
values in the FIFO queues, are set 0 except that we initialize
$\sigma_j^2=1$ for each $j$ to avoid division by 0.

\begin{figure}[t]
 \begin{minipage}{0.33\linewidth}
  \centering
  \includegraphics[width=\linewidth]{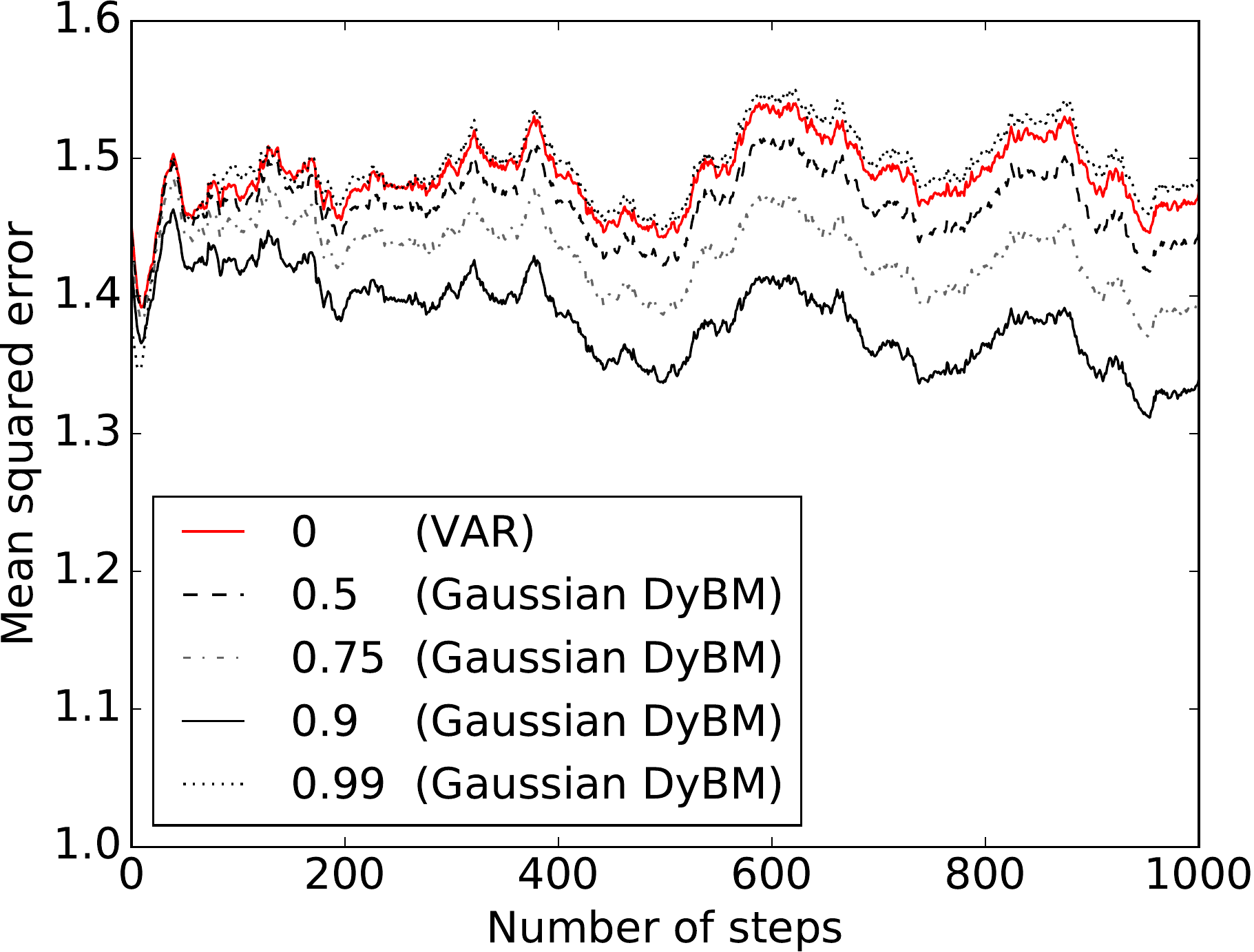}\\
  (a) Error ($d=1$)
 \end{minipage}
 \begin{minipage}{0.33\linewidth}
  \centering
  \includegraphics[width=\linewidth]{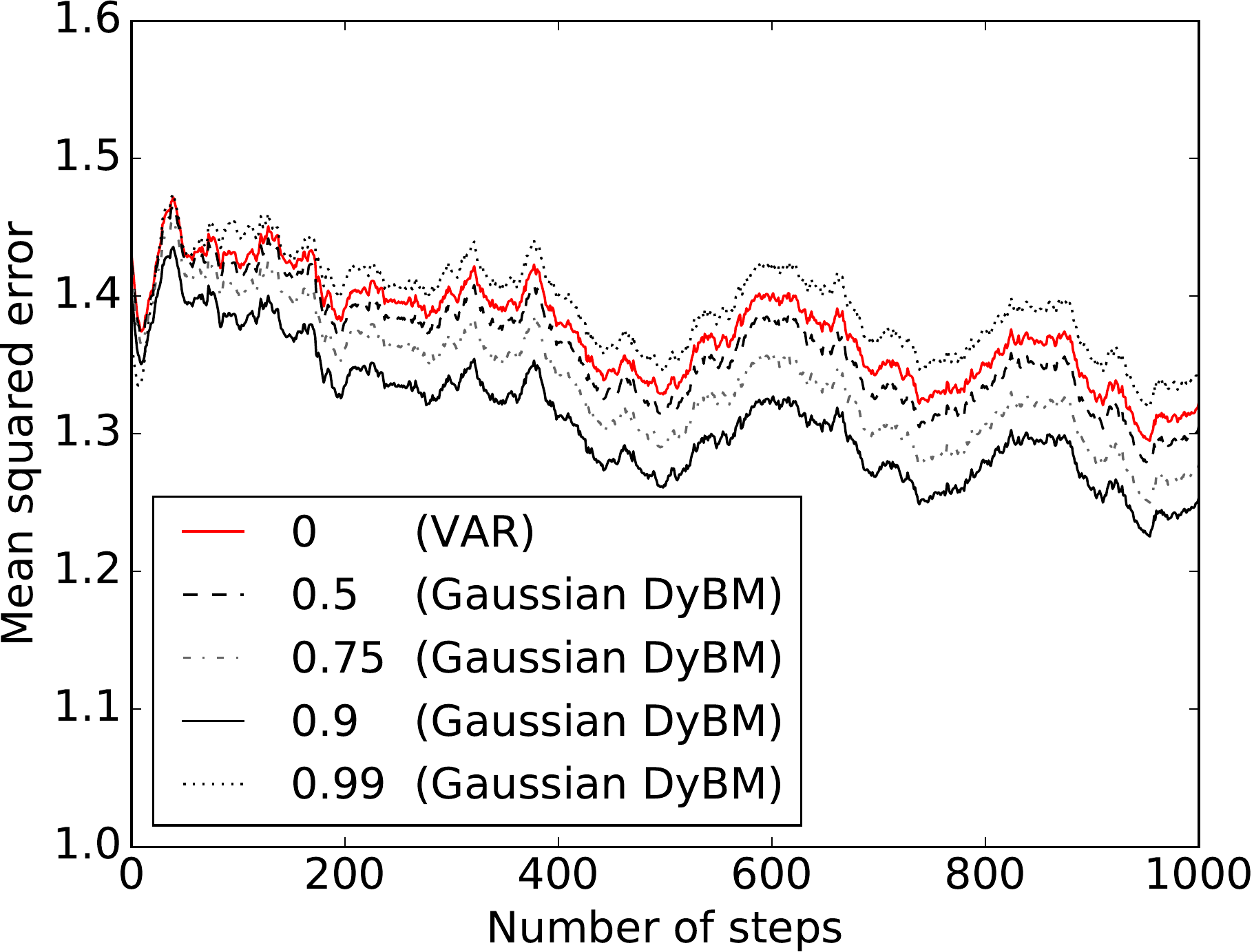}\\
  (b) Error ($d=16$)
 \end{minipage}
 \begin{minipage}{0.33\linewidth}
  \centering
  \includegraphics[width=\linewidth]{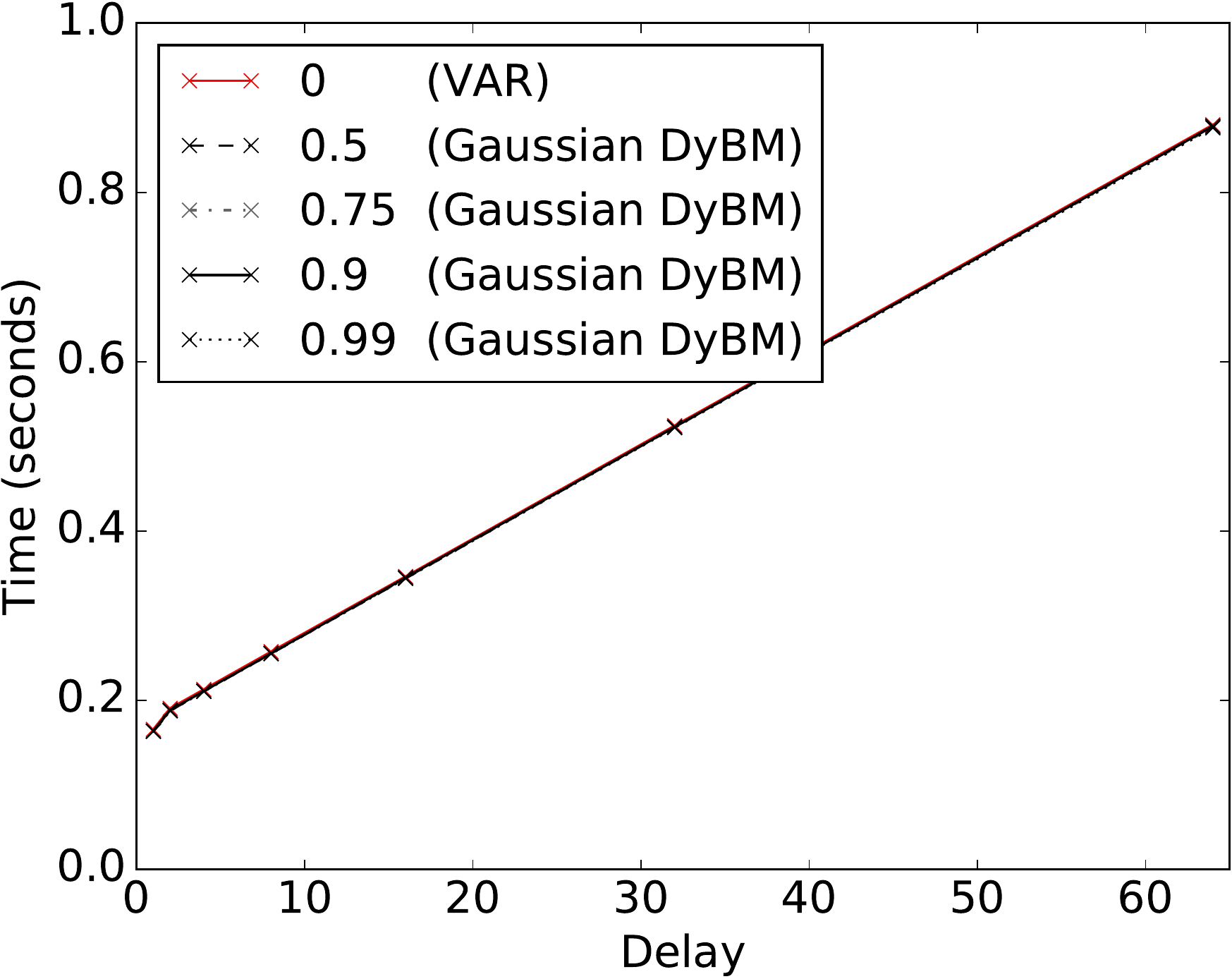}\\
  (c) Time
 \end{minipage}
 \caption{Squared error of prediction and learning time of Gaussian DyBMs
 and VAR models.  (a)-(b) For each step $t$, the squared
 error is averaged over 100 independent runs.  Decay rate $\mu$ is
 varied as in the legend, and the red curve ($\mu=0$) corresponds to
 a VAR model.  Conduction delay $d$ is varied across panels.  (c) The
 learning time per a run of 1,000 steps is plotted against delay $d$.
 Gaussian DyBMs and VAR are indistinguishable in the figure.}
 \label{fig:sin}
\end{figure}

Figure~\ref{fig:sin} shows the predictive error of the Gaussian DyBM.
Here, the prediction, $m^{[t]}$, for the pattern at time $t$ is
evaluated with mean squared error, ${\rm MSE}^{[t]}\equiv\frac{1}{100}
\sum_{s=t-50}^{50} (m^{[t]} - x^{[t]})^2$, and ${\rm MSE}^{[t]}$
is further averaged over 100 independent runs of the experiment to make
the curves in the figure smooth.  Due to the noise $\varepsilon^{[t]}$,
the best possible squared error is 1.0 in expectation.  We vary
$\mu$ as indicated in the legend and $d$ as indicated below each
panel.

Although the accuracy of the prediction with the Gaussian DyBM depends
on the choice of $\mu$, the figure shows that the Gaussian DyBM
(with $\mu>0$; black curves) generally outperforms the corresponding VAR
model ($\mu=0$; red curves) and reduces the error by up to 20~\%.
The gain that the Gaussian DyBM has over the VAR stems solely from the
use of the eligibility trace, $\gamma^{[t-1]}$, instead of the lag-$d$
variable, $x^{[t-d]}$.  
The results for longer conduction
delay can be found in Appendix~\ref{sec:additional}.

A Gaussian DyBM performs relatively well even with $d=1$, because,
unlike VAR, history beyond $d=1$ is taken into account in eligibility
traces. The figure shows that VAR performs significantly better with
$d=16$ and becomes comparable to the Gaussian DyBM with $d=1$.  A
larger $d$, however, comes at the expense of increased computational
complexity.  Here, notice that a Gaussian DyBM has essentially
equivalent computational complexity as the corresponding VAR, as we
use a single decay rate ($L=1$).  Figure~\ref{fig:sin}(c) indeed shows
that the Gaussian DyBM runs as fast as the VAR, and their learning
time grows linearly with respect to the delay $d$.  In general, for a
densely connected Gaussian DyBM, per-step computational complexity is
$O((L+d)\,N^2)$, where $L$ is the number of decay rates, $d$ is the
maximum conduction delay, and $N$ is the number of neurons. The
computational complexity is reduced to $O((L+d)\,M\,N)$ when each
neuron is connected to at most $M$ neurons.

\section{Conclusion}

A DyBM is a model of a spiking neural network, and we have shown how the 
DyBM can be used to learn binary or real-valued time-series.  For 
binary-valued time-series, the DyBM can be seen as logistic regression 
for predicting the next (spiking) pattern on the basis of the history of 
(spiking) patterns. The DyBM deals with real-valued time-series in the 
form of a Gaussian DyBM, and we have seen that the Gaussian DyBM extends 
a VAR model by including eligibility traces as additional explanatory 
variables, which allow the Gaussian DyBM to capture long term dependency 
of time-series.  Our experimental results demonstrate the effectiveness 
of the eligibility traces in increasing the predictive accuracy.

The Gaussian DyBM is only one way to deal with real values by a DyBM. In 
particular, a DyBM may assume the distribution in the exponential family 
\cite{WRH04} instead of the Gaussian distribution. The Gaussian DyBM may 
also be extended to allow nonlinear hidden units.  In \cite{RNNDyBM}, we 
will extend this preliminary manuscript and study a Gaussian DyBM with 
such extensions.

\subsubsection*{Acknowledgments}
This research is supported by CREST, JST.

\bibliographystyle{abbrv}
\bibliography{dybm}

\begin{thebibliography}{10}

\bibitem{AHS85}
D.~H. Ackley, G.~E. Hinton, and T.~J. Sejnowski.
\newblock A learning algorithm for {B}oltzmann machines.
\newblock {\em Cognitive Science}, 9:147–169, 1985.

\bibitem{NaturalGradient}
S.~Amari and H.~Nagaoka.
\newblock {\em Methods of Information Geometry}.
\newblock Oxford University Press, 2000.

\bibitem{BMFZW16}
Y.~Bengio, T.~Mesnard, A.~Fischer, S.~Zhang, and Y.~Wu.
\newblock {STDP} as presynaptic activity times rate of change of postsynaptic
  activity.
\newblock arXiv:1509.05936v2, 2016.

\bibitem{BiPoo98}
G.~Bi and M.~Poo.
\newblock Synaptic modifications in cultured hippocampal neurons: {D}ependence
  on spike timing, synaptic strength, and postsynaptic cell type.
\newblock {\em Journal of Neuroscience}, 18:10464–10472, 1998.

\bibitem{RNNDyBM}
S.~Dasgupta and T.~Osogami.
\newblock Nonlinear dynamic {B}oltzmann machines for time series prediction.
\newblock In {\em Proceedings of the Thirty-First AAAI Conference on Artificial
  Intelligence (AAAI-17)}, 2017.

\bibitem{DelayPruning}
S.~Dasgupta, T.~Yoshizumi, and T.~Osogami.
\newblock Regularized dynamic {B}oltzmann machine with delay pruning for
  unsupervised learning of temporal sequences.
\newblock In {\em Proceedings of the 23rd International Conference on Pattern
  Recognition}, 2016.

\bibitem{AdaGrad}
J.~Duchi, E.~Hazan, and Y.~Singer.
\newblock Adaptive subgradient methods for online learning and stochastic
  optimization.
\newblock {\em Journal of Machine Learning Research}, 12:2121--2159, 2011.

\bibitem{Hebb}
D.~O. Hebb.
\newblock {\em The organization of behavior: {A} neuropsychological approach}.
\newblock Wiley, 1949.

\bibitem{HinSal06}
G.~E. Hinton and R.~Salakhutdinov.
\newblock Reducing the dimensionality of data with neural networks.
\newblock {\em Science}, 313:504–507, 2006.

\bibitem{HinSej83}
G.~E. Hinton and T.~J. Sejnowski.
\newblock Optimal perceptual inference.
\newblock In {\em Proc. IEEE Conference on Computer Vision and Pattern
  Recognition}, pages 448--453, June 1983.

\bibitem{GaussianBM}
T.~Marks and J.~Movellan.
\newblock Diffusion networks, products of experts, and factor analysis.
\newblock In {\em Proceedings of the Third International Conference on
  Independent Component Analysis and Blind Source Separation}, 2001.

\bibitem{NPBM13}
B.~Nessler, M.~Pfeiffer, L.~Buesing, and W.~Maass.
\newblock Bayesian computation emerges in generic cortical microcircuits
  through spike-timing-dependent plasticity.
\newblock {\em PLoS Computational Biology}, 9(4):e1003037, 2013.

\bibitem{HyperDyBM}
T.~Osogami and S.~Dasgupta.
\newblock Learning the values of the hyperparameters of a dynamic {B}oltzmann
  machine.
\newblock {\em IBM Journal of Research and Development}, 61(4/5):to appear,
  2017.

\bibitem{RT0967}
T.~Osogami and M.~Otsuka.
\newblock Learning dynamic {B}oltzmann machines with spike-timing dependent
  plasticity.
\newblock Technical Report RT0967, IBM Research, 2015.

\bibitem{DyBM}
T.~Osogami and M.~Otsuka.
\newblock Seven neurons memorizing sequences of alphabetical images via
  spike-timing dependent plasticity.
\newblock {\em Scientific Reports}, 5:14149, 2015.

\bibitem{SceBen16}
B.~Scellier and Y.~Bengio.
\newblock Equilibrium propagation: Bridging the gap between energy-based models
  and backpropagation.
\newblock arXiv:1602.05179v4, 2016.

\bibitem{WRH04}
M.~Welling, M.~Rosen-{Z}vi, and G.~E. Hinton.
\newblock Exponential family harmoniums with an application to information
  retrieval.
\newblock In {\em Advances in Neural Information Processing Systems 17}, pages
  1481--1488. MIT Press, 2004.

\end{thebibliography}

\appendix
\section{Supplementary material for \textit{Gaussian dynamic Boltzmann machines}}

\subsection{Natural gradients}
\label{sec:natural}

Consider a stochastic model that gives the probability density
of a pattern $\mathbf{x}$ as $p(\mathbf{x};\theta)$.
With natural gradients \cite{NaturalGradient}, the parameters,
$\theta$, of the stochastic model are updated as follows:
\begin{align}
 \theta_{t+1} = \theta_t - \eta_t \, G^{-1}(\theta_t) \, \nabla \log p(\mathbf{x};\theta)
 \label{eq:natural}
\end{align}
at each step $t$, where $\eta_t$ is the learning rate at $t$, and
$G(\theta)$ denotes the Fisher information matrix:
\begin{align}
 G(\theta) & \equiv \int p(\mathbf{x};\theta) \left(\nabla \log p(\mathbf{x};\theta) \, \nabla \log p(\mathbf{x};\theta)^\top \right) d\mathbf{x}.
\end{align}

Due to the conditional independence in \eqref{eq:independence}, it
suffices to
derive a natural gradient for each Gaussian unit.  Here, we consider the
parametrization with mean $m$ and variance $v\equiv\sigma^2$.  The
probability density function of a Gaussian distribution is represented
with this parametrization as follows:
\begin{align}
p(x;m,v) & = \frac{1}{\sqrt{2\pi\,v}} \exp\left(-\frac{(x-m)^2}{2v}\right).
\end{align}
The log likelihood of $x$ is then given by
\begin{align}
 \log p(x;m,v) & = -\frac{(x-m)^2}{2v} - \frac{1}{2} \log v - \frac{1}{2}\log 2\pi.
 \label{eq:loglikelihood}
\end{align}

Hence, the gradient and the inverse Fisher information matrix in \eqref{eq:natural} are given as follows:
\begin{align}
 \nabla \log p(\mathbf{x};\theta) & = \left(\begin{array}{c}
		       \frac{x-m}{v}\\
			    \frac{(x-m)^2}{2v^2} - \frac{1}{2v}
					    \end{array}\right)\\
 G^{-1}(\theta)  &
 = \left(\begin{array}{cc}
    \frac{1}{v} & 0 \\
    0 & \frac{1}{2v^2}
   \end{array}\right)^{-1}
 = \left(\begin{array}{cc}
    v & 0 \\
    0 & 2v^2
   \end{array}\right),
\end{align}

The parameters $\theta_t\equiv(m_t,v_t)$ are then updated as follows:
\begin{align}
m_{t+1}
 & = m_t + \eta_t \, (x-m_t) \\
v_{t+1}
 & = v_t + \eta_t \, \left((x-m_t)^2 - v_t\right).
\end{align}

In the context of a Gaussian DyBM, the mean is given by \eqref{eq:mean},
where $m_j^{[t]}$ is linear with respect to $b_j$, $w_{i,j}$, and $u_{i,j,\ell}$.
Also, the variance is given by $\sigma_j^2$.  Hence, the natural gradient gives
the learning rules for these parameters as follows:
\begin{align}
 b_j \leftarrow b_j + \eta \, \big(x_j^{[t]}-m_j^{[t]}),
  & \hspace{10mm}
\sigma_j^2 \leftarrow \sigma_j^2 + \eta \, \big( (x_j^{[t]}-m_j^{[t]})^2 - \sigma_j^2 \big),
 \label{eq:natural_rule1}
 \\
 w_{i,j}^{[\delta]} \leftarrow w_{i,j}^{[\delta]} + \eta \, \big(x_j^{[t]}-m_j^{[t]}) \, x_i^{[t-\delta]},
 & \hspace{10mm}
 u_{i,j,\ell} \leftarrow u_{i,j,\ell} + \eta \, \big(x_j^{[t]}-m_j^{[t]})\, \alpha_{i,\lambda_{\ell}}^{[t-1]},
 \label{eq:natural_rule2}
\end{align}
which can be compared against what the standard gradient gives in
\eqref{eq:learning_rule1}-\eqref{eq:learning_rule2}.

\subsection{Additional results of experiments}
\label{sec:additional}

\begin{figure}[b]
  \begin{minipage}{0.5\linewidth}
  \centering
  \includegraphics[width=0.8\linewidth]{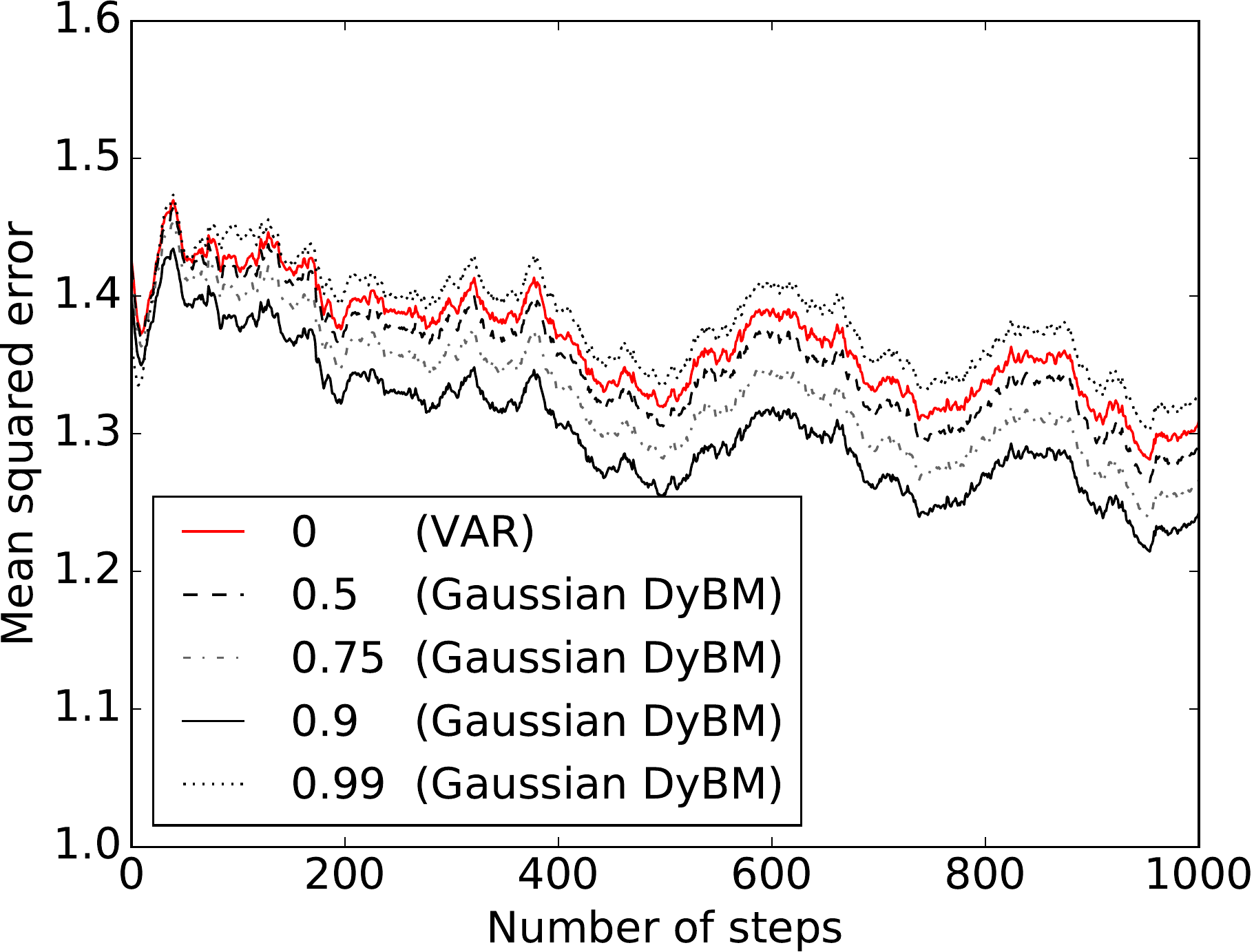}\\
  (a) $d=32$
  \end{minipage}
  \begin{minipage}{0.5\linewidth}
    \centering
  \includegraphics[width=0.8\linewidth]{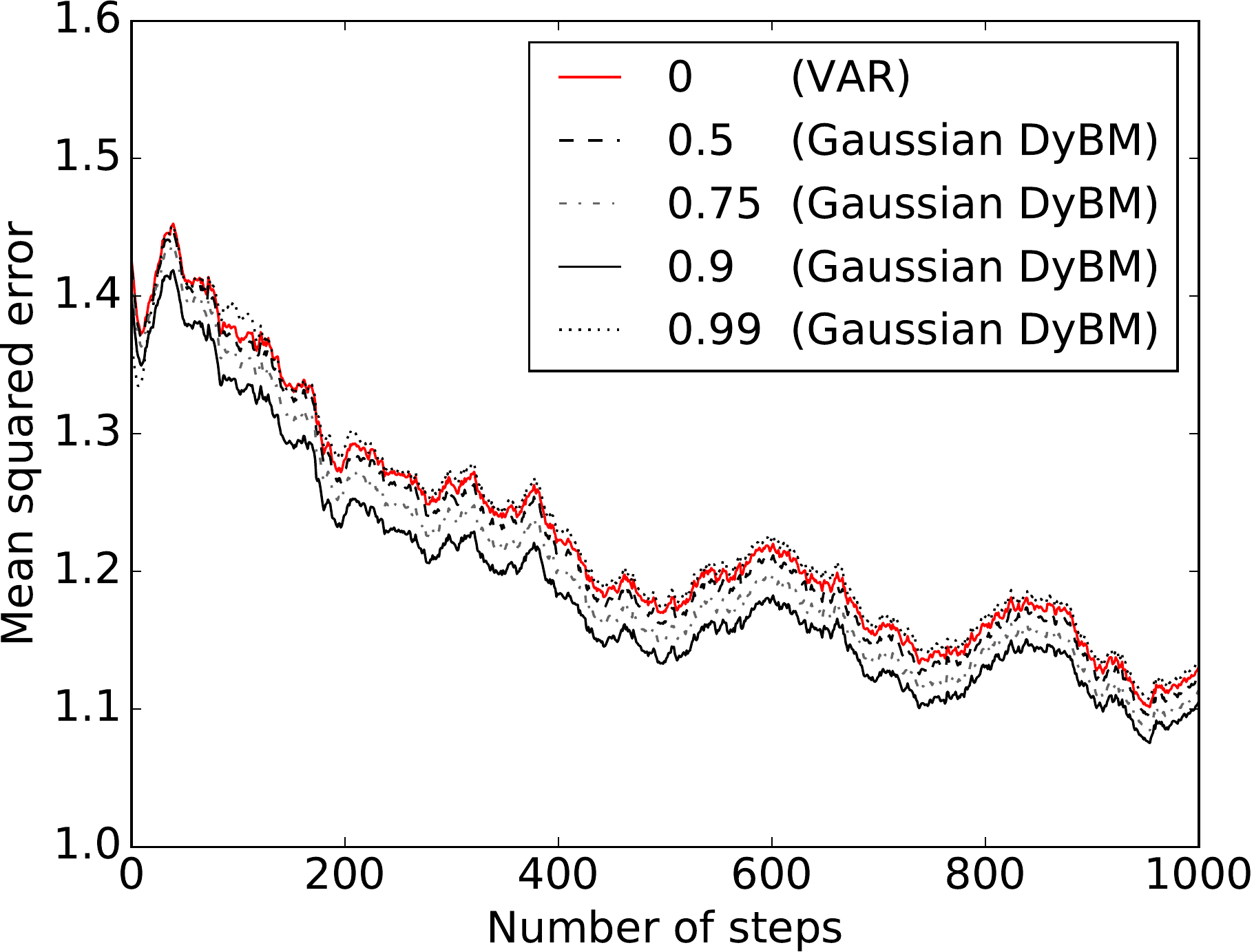}\\
  (b) $d=64$
  \end{minipage}
 \caption{The results with longer conduction delay $d$ for the experiments in Figure~\ref{fig:sin}.}
 \label{fig:sin2}
\end{figure}

Figure~\ref{fig:sin2} shows additional results of the experiments shown
in Figure~\ref{fig:sin}.  Now, the conduction delay varies from $d=32$
to $d=64$.  Learning the noisy sine wave \eqref{eq:sin} becomes rather
trivial with $d>50$, because the expected value of the noisy sine wave
with the period of 100 satisfies ${\sf E}[x^{[t]}] = - {\sf
  E}[x^{[t-50]}]$.  Namely, \ref{eq:1Dmodel} can exactly represent
this noisy sine wave by setting $w^{[50]}=1$ and other parameters zero.

\end{document}